\documentclass[10pt,twocolumn,letterpaper]{article}

\usepackage{iccv}
\usepackage{times}
\usepackage{epsfig}
\usepackage{graphicx}
\usepackage{amsmath}
\usepackage{amssymb}

\newcommand*\samethanks[1][\value{footnote}]{\footnotemark[#1]}

\usepackage[pagebackref=true,breaklinks=true,letterpaper=true,colorlinks,bookmarks=false]{hyperref}

\iccvfinalcopy %

\def\iccvPaperID{6331} %

\ificcvfinal\fi
\begin{document}

\title{Few-Shot Video Classification via Temporal Alignment}

\author{Kaidi Cao \hspace{9pt} Jingwei Ji\thanks{Indicates equal contribution.} \hspace{9pt} Zhangjie Cao\samethanks \hspace{9pt} Chien-Yi Chang \hspace{9pt} Juan Carlos Niebles\\
Stanford University\\
{\tt\small \{kaidicao, jingweij, caozj18, cy3, jniebles\}@cs.stanford.edu}
}

\maketitle

\begin{abstract}
There is a growing interest in learning a model which could recognize novel classes with only a few labeled examples. 
In this paper, we propose Temporal Alignment Module (TAM), a novel few-shot learning framework that can learn to classify a previous unseen video. While most previous works neglect long-term temporal ordering information, our proposed model explicitly leverages the temporal ordering information in video data through temporal alignment. This leads to strong data-efficiency for few-shot learning. In concrete, TAM calculates the distance value of query video with respect to novel class proxies by averaging the per frame distances along its alignment path. We introduce continuous relaxation to TAM so the model can be learned in an end-to-end fashion to directly optimize the few-shot learning objective. We evaluate TAM on two challenging real-world datasets, Kinetics and Something-Something-V2, and show that our model leads to significant improvement of few-shot video classification over a wide range of competitive baselines.

\end{abstract}

\section{Introduction}

\begin{figure}[t]
    \centering
    \includegraphics[width=0.85\linewidth]{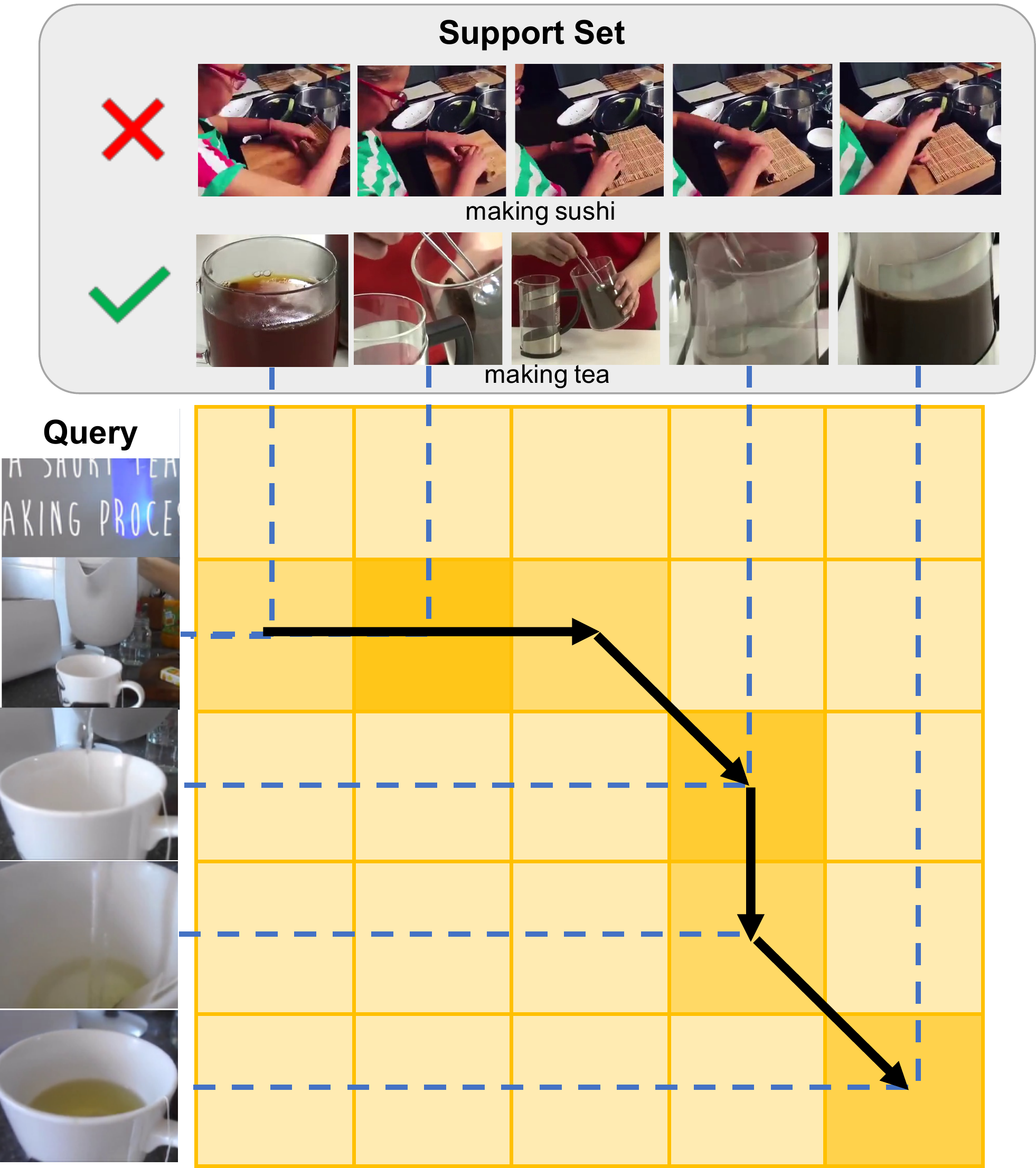}
    \caption{\textbf{Our few-shot video classification setting}. %
    Pairs of semantically matched frames are connected with a blue dashed line. The arrows show the direction of the temporal alignment path.}
    \label{fig:teaser}
\end{figure}

The emergence of deep learning has greatly advanced the frontiers of action recognition \cite{wang2016temporal,carreira2017quo}. The main focus tends to center around learning effective video representations for classification using large amounts of labeled data.  In order to recognize novel classes that a pretrained network has not seen before, typically we need to manually collect hundreds of video samples for knowledge transferring. But such a procedure is rather tedious and labor intensive especially for videos, where the difficulty and cost of labeling is much higher compared to images.

There is a growing interest in learning models capable of effectively adapting themselves to recognize novel classes with only a few labeled examples. This is known as the few-shot learning task \cite{garcia2017few,chen19closerfewshot}. Under the setup of meta-learning based few-shot learning, the model is explicitly trained to deal with scarce training data for previously unseen classes across different episodes. While the majority of recent few-shot learning works focus on image classification, adapting it to video data is not a trivial extension.

Videos are much more complicated than images, as recognizing some specific actions, such as opening the door, usually requires a complete modeling of temporal information. In the previous literature of video classification, 3D convolution and optical flow are two of the most popular methods to model temporal relations. The direct output of neural network encoders is always a temporal sequence of deep encoded features.
State-of-the-art approaches commonly apply a temporal pooling module (usually mean pooling) in order to make final prediction. As observed before, averaging the deep features only captures the co-occurrence rather than the temporal ordering of patterns, which will unavoidably result in information loss. 

Loss of information is even more severe for few-shot learning. 
It is hard to learn the local temporal patterns which are useful for few-shot classification with limited amount of data. Utilizing the long-term temporal ordering information, which is often neglected in previous works on video classification, might potentially help with few-shot learning. For example, if the model could verify that there is a procedure of pouring water before a close-up view of a just made tea, as shown in Fig. \ref{fig:teaser}, the model will then become quite confident about predicting the class of this query video to be making tea, rather than some other potential predictions like boiling water or serving tea. 
In addition, Fig.~\ref{fig:teaser} also shows that for two videos in the same class, even though they both contain a procedure of pouring water followed by closed-up view of tea, the exact temporal duration of each atomic step can vary dramatically. This non-linear temporal variations of videos pose great challenges for few-shot video learning. 

With these insights, we thus propose Temporal Alignment Module (TAM) for few-shot video classification, a novel temporal-alignment based approach that learns to estimate temporal alignment score of a query video with corresponding proxies in the support set. To be specific, we compute temporal alignment score for each potential query-support pair by averaging per-frame distances along a temporal alignment path, which enforces the score we use to make prediction to preserve temporal ordering. Furthermore, TAM is fully differentiable so that the model can be trained end-to-end and optimize the few-shot objective directly. This in turn helps the model to better utilize long-term temporal information to make few-shot learning predictions. This module allows us to better model the temporal evolution of videos, while enabling stronger data efficiency in the few-shot setting. 

We evaluate our model for few-shot video classification task on two action recognition datasets: Kinetics \cite{kay2017kinetics} and Something-Something V2 \cite{goyal2017something}. We show that when there is only a single example available, our method outperforms the mean pooling baseline which does not consider temporal ordering information by approximately 8\% in top-1 accuracy. We also show qualitatively that the proposed framework is able to learn meaningful alignment path in an end-to-end manner.

In summary, our main contributions are: (i) We are the first to explicitly address the non-linear temporal variations issue in the few-shot video classification setting. (ii) We propose Temporal Alignment Module (TAM), a data-efficient few-shot learning framework that can dynamically align two video sequences while preserving the temporal ordering, which is often neglected in previous works. (iii) We use continuous relaxation to make our model fully differentiable and show that it outperforms previous state-of-the-art methods by a large margin on two challenging datasets.

\section{Related Work}

\noindent \textbf{Few-Shot Learning.}
To address few-shot learning, a direct approach is to train a model on the training set and fine-tune with the few data in the novel classes. Since the data in novel classes are not enough to fine-tune the model with general learning techniques, methods are proposed to learn a good initialization model \cite{finn2017model,nichol2018reptile,rusu2018meta} or develop a novel optimizer \cite{ravi2016optimization,munkhdalai2017meta}. These works aim to relieve the difficulty of fine-tuning the model with limited samples. However, such methods suffer from overfitting when the training data in novel classes are scarce but the variance is large. Another branch of works, which learns a common metric for both seen and novel classes, can avoid overfitting to some extent. Convolutional Siamese Net \cite{koch2015siamese} trains a Siamese network to compare two samples. Latent Embedding Optimization \cite{vinyals2016matching} employs attention
kernel to measure the distance. Prototypical Network \cite{snell2017prototypical} utilizes the Euclidean distance to the class center. Graph Neural Networks \cite{garcia2017few} constructs a weighted graph to represent all the data and measure the similarity between data. Other methods use data augmentation, which learns to augment labeled data in unseen classes for supervised training \cite{hariharan2017low,wang2018low}. However, video generation is still an under-explored problem at least generating videos condition on a typical category. Thus, in this paper, we employ the metric learning approach and designs a temporal-aligned video metric for few-shot video classification.

There are works exploring few-shot recognition. OSS-Metric Learning \cite{kliper2011one} proposes a novel OSS-Metric Learning to measure the similarity of video pairs to enable one-shot video classification. \cite{mishra2018generative} introduces a zero-shot method which learns a mapping function from an attribute to a class center. It has an extension to few-shot learning by integrating labeled data on unseen classes. CMN \cite{zhu2018compound} is the most related work to ours. They introduce a multi-saliency embedding algorithm to encode video into a fixed-size matrix representation. Then they propose a compound memory network (CMN) to compress and store the representation and classify videos by matching and ranking. However, previous works collapse the order of frames at representation \cite{kliper2011one,mishra2018generative,zhu2018compound}. Thus, the learned model is sub-optimal for video datasets where sequence order is important. In this paper, we preserve the frame order in video representation and estimate distance with temporal alignment, which utilizes video sequence order to solve few-shot video classification.

\begin{figure*}[t]
    \centering
    \includegraphics[width=0.9\textwidth]{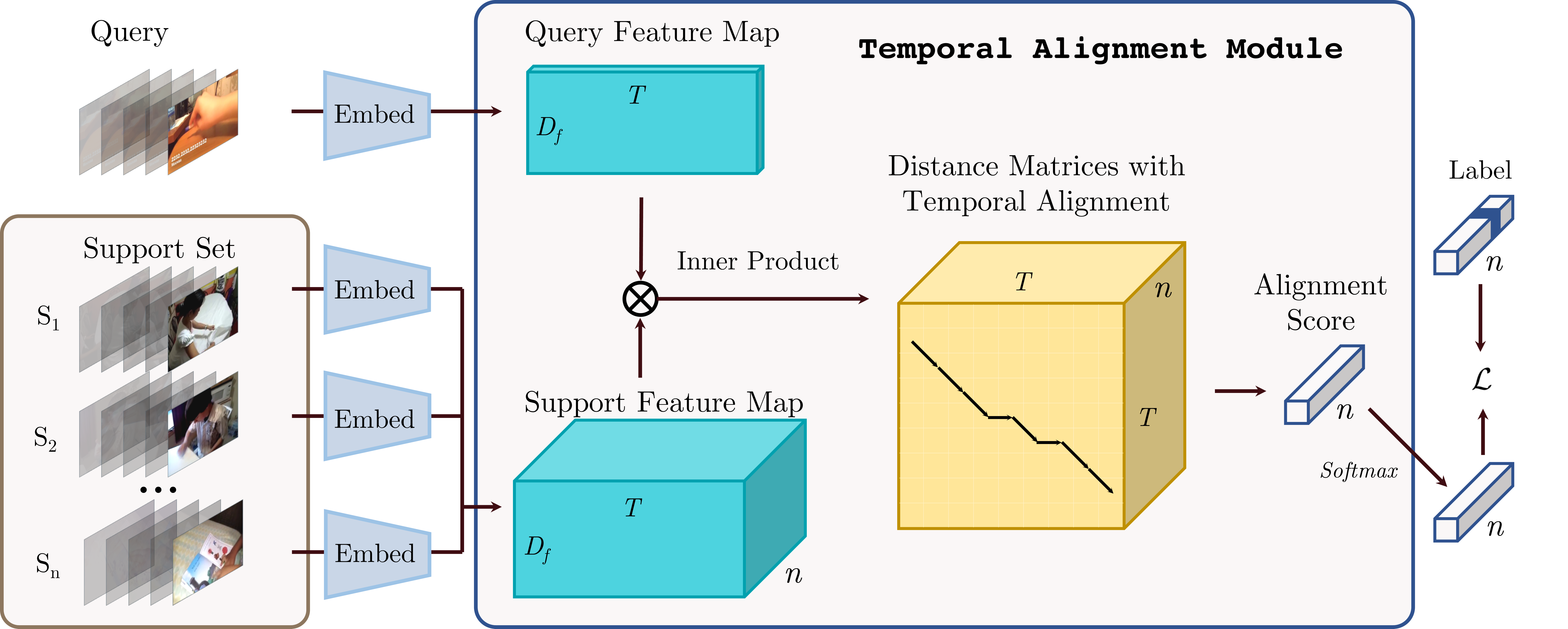}
    \caption{\textbf{Overview of our method}. We first extract per-frame deep features using the embedding network. We then compute the distance matrices between the query video and videos in the support set. Next, an alignment score is computed out of the matrix representation. Finally we apply softmax operator over the alignment score of each novel class.} 
    \label{fig:pipeline}
\end{figure*}

\noindent \textbf{Video Classification.}
A significant amount of research has tackled the problem of video classification. State-of-the-art video classification methods have evolved from hand-crafted representation learning \cite{3dhog, sift3d, idt} to deep-learning based models. C3D \cite{tran2015learning} utilizes 3D spatial-temporal convolutional filters to extract deep features from sequences of RGB frames. TSN \cite{wang2016temporal} and I3D \cite{carreira2017quo} uses two-stream 2D or 3D CNNs with larger size on both RGB and optical flow sequences. By factorizing 3D convolutional filters into separate spatial and temporal components, P3D \cite{p3d} and R(2+1)D \cite{r2+1d} yield models with comparable or superior classification accuracy but smaller in size. An issue of these video representation learning methods is their dependence on large-scale video datasets for training. Models with an excessive amount of learnable parameters tend to fail when only a small number of training samples are available.

Another concern of video representation learning is the lack of temporal relational reasoning. Classification on the videos sensitive to temporal ordering poses a more significant challenge to the above networks which are tailored to capture short-term temporal features. Non-local neural networks \cite{wang2018non} introduce self-attention to aggregate temporal information in the long-term. Wang \etal \cite{wang2018videos} further employ space-time region graphs to model the spatial-temporal reasoning. Recently, TRN \cite{zhou2018temporal} proposes a temporal relational module to achieve superior performance. Still, these networks inevitably pool/fuse features from different frames in the last layers to extract a single feature vector representing the whole video. In contrast, our model is able to learn video representation without loss of temporal ordering in order to generate more accurate final predictions.

\noindent \textbf{Sequence Alignment.}
Sequence alignment is of great importance in the field of bioinformatics, which describes the way of arrangement of DNA/RNA or protein sequences, in order to identify the regions of similarity among them~\cite{altschul1997gapped}. In the vision community, researchers have growing interests in tackling the sequence alignment problem with high dimensional multi-modal data, such as finding the alignment between untrimmed video sequence and the corresponding textual action sequence~\cite{chang2019d, dogan2018neural,richard2018neuralnetwork}. The main technique that has been applied to this line of work is dynamic programming. While dynamic programming is guaranteed to find the optimal alignment between two sequences given a prescribed distance function, the discrete operations used in dynamic programming are non-differentiable and hence prevent learning distance functions with gradient-based methods. Our work is closely related to recent progress on using continuous relaxation of discrete operations to tackle sequence alignment problem~\cite{chang2019d} and hence allow us to train our entire model end-to-end.

\section{Methods}

Our goal is to learn a model which can classify novel classes of videos with only a few labeled examples. The wide range of intra-class spatial-temporal variations of videos poses great challenges for few-shot video classifications. We address this challenge by proposing a few-shot learning framework with Temporal Alignment Module (TAM), which is to our best knowledge the first model that can explicitly learn a distance measure independent of non-linear temporal variations in videos. The use of TAM sets our approach apart from previous works that fail to preserve temporal ordering and relation during meta training and meta testing. Fig.\ref{fig:pipeline} shows the outline of our model. 

In the following, we will first provide a problem formulation of few-shot video classification task, and then define our model and show how it can be used at training and test time. 

\subsection{Problem Formulation}

In the few-shot video classification setting, we split the classes we have annotation of into $\mathcal{C}_{train}$: the base classes that have sufficient data for representation learning and $\mathcal{C}_{test}$: the novel or unseen classes that have only a few labeled data during testing stage. The goal of few-shot learning is then to train a network that can generalize well to new episodes over novel classes. In a $n$-way, $k$-shot problem, for each episode the support set will contain $n$ novel classes, and each class will have a very small amount of samples ($k$ in our setting). The algorithm will have to classify videos from query set to one of the novel classes in support set. Episodes are randomly drawn from a larger collection of data, which we hereby denote them as meta set. In our setting, we introduce 3 splits over classes as meta training $\mathcal{T}_{train}$, meta validation $\mathcal{T}_{val}$ and meta testing $\mathcal{T}_{test}$ sets. 

We formulate the few-shot learning as a representation learning problem through a distance function $\phi(f_{\varphi}(x_1), f_{\varphi}(x_2))$, where $x_1$ and $x_2$ are two samples drawn from $\mathcal{C}_{train}$ and $f_{\varphi}(\cdot)$ is an embedding function that maps samples to their representations. The difference between our problem formulation with the majority of previous few-shot learning researches lies in the fact that we are now dealing with higher dimensional inputs, i.e. (2+1)D volumes instead of 2D images. The addition of the time dimension in few-shot setting demands the model to be able to learn temporal ordering and relation with limited data in order to generalize to novel classes, which pose challenges that have not been properly addressed by previous works.

\subsection{Model}
With the above problem formulation, our goal is to learn a video distance function by minimizing the few-shot learning objective. Our key insight is that we want to explicitly learn a distance function independent of non-linear temporal variations by aligning the frames of two videos. Unlike previous works which use weighted average or mean pooling along the time dimension~\cite{wang2016temporal,tran2015learning,wang2018non,xie2018rethinking,carreira2017quo,zhu2018compound}, our model is able to infer temporal ordering and relationship during meta training or meta testing in an explicit and data efficient manner. In this subsection, we will breakdown our model following the pipeline as illustrated in Fig.~\ref{fig:pipeline}.

\noindent
\textbf{Embedding Module:} The purpose of the embedding module $f_{\varphi}$ is to generate a compact representation of a trimmed action video that encapsulates its visual content. A raw video usually consists of hundreds of frames, whose information could be redundant were to perform per frame inference. Thus frame sampling is usually adopted as a preproccessing stage for video inputs. Existing frame sampling schemes can be mainly divided into two categories: dense sampling (randomly cropping out
$T$ consecutive frames from the original full-length video and
then dropping every other frame) \cite{tran2015learning, carreira2017quo, wang2018non, xie2018rethinking} and sparse sampling (samples distribute uniformly along the temporal dimension) \cite{wang2016temporal, zolfaghari2018eco, zhou2018temporal, lin2018temporal}. We follow the sparse sampling protocol first described in TSN \cite{wang2016temporal}, which divides the video sequence into $T$ segments and extracts a short snippets in each segment. The sparse sampling scheme allows each video sequence to be represented by a fix number of snippets. The sampled snippets span the whole video, which enables long-term temporal modeling.

Given a input sequence $S = \{ X_1, X_2, ..., X_T \}$, we will apply a CNN backbone network $f_{\varphi}$ to process each of the snippets individually. After the encoding, raw video snippets will then turn into a sequence of feature vectors $f_{\varphi}(S) = \{f_{\varphi}(X_1), f_{\varphi}(X_2), ..., f_{\varphi}(X_T)\}$. It is worth noticing that for the embedding of each video $f_{\varphi}(S)$, its dimension is $T \times D_f$, rather than $D_f$ for image embedding, which is usually chosen as the activation before final fully-connected layer of a CNN network.

\noindent
\textbf{Distance Measure with Temporal Alignment Module (TAM):} 

Given two videos $S_i, S_j$ and their embedded features $f_{\varphi}(S_i), f_{\varphi}(S_j) \in \mathbb{R}^{T \times D}$, we can calculate the frame-level distance matrix $D \in \mathbb{R}^{T \times T}$ as 
\begin{align}
    D_{l,m} = 1 - \frac{f_{\varphi}(S_i)_{l,} \cdot f_{\varphi}(S_j)_{m,}}{\lvert\lvert f_{\varphi}(S_i)_{l,}\rvert \rvert \; \lvert\lvert f_{\varphi}(S_j)_{m,}\rvert \rvert},
\label{eq:similarity}
\end{align}
where $D_{l,m}$ is the frame-level distance value between the $l$th frame of video $S_i$ and the $m$th frame of video $S_j$. 
\begin{figure*}[h]
    \centering
    \includegraphics[width=0.9\textwidth]{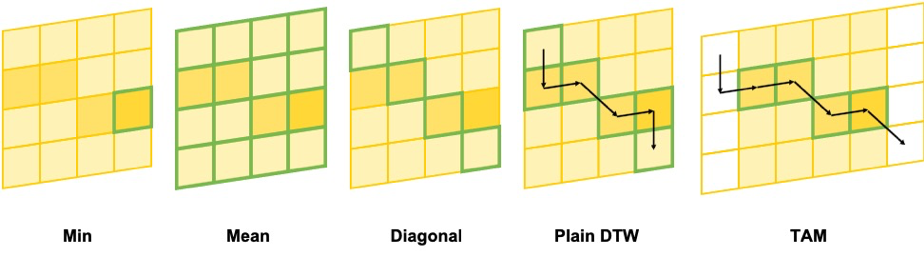}
    \caption{\textbf{Methods for calculating alignment score}. Each subplot shows a distance matrix. The darker of the color of an entry, the smaller the distance value is of a pair of relevant frames. The entries with green border denotes the entries contributing to the final alignment score.}
    \label{fig:DTW}
\end{figure*}

We further define $\mathcal{W} \subset \{0,1\}^{T\times T}$ to be the set of possible binary alignment matrices, where $\forall W \in \mathcal{W}$, $W_{ij} = 1$ if the $i$th frame of video $S_i$ is aligned to the $j$th frame of video $S_j$. Our goal is to find the best alignment $W^* \in \mathcal{W}$
\begin{align}
W^{*} = \underset{W \in \mathcal{W}}{\mathrm{argmin}} \langle W, D(f_{\varphi}(S_i),f_{\varphi}(S_j))\rangle,
\label{eqn:dtw_objective}
\end{align}
which minimizes the inner product between the alignment matrix $W$ and the frame-level distance matrix $D$ defined in Eq.~\eqref{eq:similarity}. The video distance measure is thus given by 
\begin{align}
\phi(f_{\varphi}(S_i), f_{\varphi}(S_j)) = \langle W^*, D \rangle.  
\label{eq:video_score}
\end{align}
We propose to use a variant of Dynamic Time Warping (DTW) algorithm~\cite{muller2007dynamic} to solve Eq.~\eqref{eqn:dtw_objective}. This is achieved by solving for a cumulative distance function
{\small \begin{align}
 \gamma(i,j) = D_{ij} + \min \{ \gamma(i-1, j-1), \gamma(i-1,j), \gamma(i,j-1) \}.
\end{align}}

In this setting of plain DTW, an alignment path is a contiguous set of matrix elements which defines a mapping between two sequences that satisfies the following conditions: boundary conditions, continuity and monotonicity. The boundary condition poses constraints on alignment matrix $W$ such that $W_{11} = 1$ and $W_{TT} = 1$ must be true in all possible alignment paths. 
In our alignment formulation, though the videos are trimmed, the action in the query video does not have to match exactly about its start and end action with the proxy. For example, consider the action of making coffee, there might be a snippet of stirring coffee at the end of action and it might not. To address this issue, we propose to relax this the boundary condition. Instead of having a path aligning the two videos from start to end, we allow the algorithm to find a path with flexible starting and ending points, while maintaining continuity and monotonicity. To work through this, we pad two column of $0$s at the start and end of the distance matrix so that it functions as enabling the alignment process to start and end at arbitrary position. So for our method, instead of computing the alignment score on a $T\times T$ matrix, we work with the padded matrix of size $T\times(T+2)$. We further denote the indexes of the first dimension as $1, 2, ..., T$, and indexes of the second dimension as $0, 1, 2, ..., T, T+1$, for simplicity. The compute of cumulative distance function is then changed into function
{\small \begin{align}
& \gamma(i, j) = \nonumber \\ 
& D_{ij} +
\begin{cases}
 \min \{ \gamma(i-1, j-1), \gamma(i-1,j), \gamma(i,j-1) \}, \\ 
 \qquad \qquad \qquad \qquad \qquad j = 0 \text{ or } j = T + 1 \\
 \min \{ \gamma(i-1, j-1), \gamma(i,j-1), \; \text{otherwise}
\end{cases}
\label{cumul distance}
\end{align}}
Note that if we follow the Eq. \ref{cumul distance} to compute the alignment score, the score by itself is naturally normalized. Since at each time step except $j=0$ and $j=T-1$, the alignment function forces a path from $\gamma(\cdot,j-1)$ to $\gamma(\cdot,j)$, the final alignment score will be a summation of exactly $T$ scores. In this light, alignment scores computed from different pairs of query videos and support videos are normalized, which means that the scale will not be affected by the path chosen.

\noindent
\textbf{Differentiable TAM with Continuous Relaxation:} While the above formulation is straightforward, the key technical challenge is that $\gamma$ is not differentiable with respect to the distance function $D$. Following the recent works on continuous relaxation of discrete operation and its application in video temporal segmentation~\cite{chang2019d, mensch2018differentiable}, we introduce a continuous relaxation to our Temporal Alignment Module (TAM). We use log-sum-exp with a smoothing parameter $\lambda > 0$ to approximate the non-differentiable minimum operator in Eq.~\eqref{cumul distance}
\begin{align}
\min (x_1, x_2, ..., x_n) \approx - \lambda \log \sum_{i=1}^n e^{-x_i/\lambda} \text{ if } \lambda \rightarrow 0.
\label{eq:relaxation}
\end{align}
While the use of continuous relaxation in Eq.~\eqref{eq:relaxation} does not convexify the the objective function, it helps the optimization process and allows gradients to be backpropagated through TAM. 

\noindent
\textbf{Training and Inference:}
We have shown how to compute the cumulative distance function $\gamma$ and use continuous relaxation to make it differentiable given a pair of input videos $(S_i, S_j)$. The video distance measure is given by
\begin{align}
    \phi(f_{\varphi}(S_i), f_{\varphi}(S_j)) = \gamma(T, T+1).
\end{align}
In training time, given ground-truth video pair $(S, \hat{S})$ and support set $\mathcal{S}$, we train our entire model end-to-end by directly minimizing the loss function
\begin{align}
& \mathcal{L} = - \log \frac{\exp(-\phi(f_{\varphi}(S), f_{\varphi}(\hat{S})))}{\sum_{Z \in \mathcal{S}} \exp(-\phi(f_{\varphi}(S), f_{\varphi}(Z)))}.
\label{eq:loss}
\end{align}
At test time we are given an unseen query video $Q$ and its support set $\mathcal{S}$, our goal is to find the video $S^* \in \mathcal{S}$ that minimize the video distance function
\begin{align}
    S^* = \underset{S \in \mathcal{S}}{\mathrm{argmin}} \; \phi(Q, S).
\end{align}

\section{Experiments}

In this work, our task is few-shot video classification, where the objective is to classify novel classes with only a few examples from the support set. We divide our experiments into the following sections. 

\subsection{Datasets}

As pointed out by \cite{xie2018rethinking,zhou2018temporal}, existing action recognition datasets can be roughly classified into two groups: Youtube type videos: UCF101 \cite{soomro2012ucf101}, Sports 1M \cite{karpathy2014large}, Kinetics \cite{kay2017kinetics}, and crowd-sourced videos: Jester\cite{jester}, Charades \cite{sigurdsson2016hollywood}, Something-Something V1\&V2 \cite{goyal2017something}, in which the videos are collected by asking the crowd-source workers to record themselves performing instructed activities. Crowd-sourced videos usually focus more on modeling the temporal relationships, since visual contents among different classes are more similar than those of Youtube type videos. To demonstrate the effectiveness of our approach on these two groups of video data, we base our few-shot evaluation on two action recognition datasets, Kinetics \cite{kay2017kinetics} and Something-Something V2 \cite{goyal2017something}. 

Kinetics \cite{kay2017kinetics} and Something-Something V2 \cite{goyal2017something} are constructed to serve as action recognition datasets so we have to construct their few-shot versions. For Kinetics dataset, we follow the same split as CMN \cite{zhu2018compound} and sample 64 classes for meta training, 12 classes for validation and 24 classes for meta testing. Since there is no existing split for few-shot classification on Something-Something V2, we construct a few-shot dataset following the same rule as CMN \cite{zhu2018compound}. We randomly selected 100 classes from the whole dataset. The 100 classes are then split into 64, 12 and 24 classes as the meta-training, meta-validation and meta-testing set, respectively.

\subsection{Implementation Details}

For a $n$-way, $k$-shot test setting, we randomly sample $n$ classes with each class containing $k$ examples as the support set. We construct the query set to have $n$ examples, where each unlabeled sample in the query set belongs to one of the $n$ classes in the support set. Thus each episode has a total of $n(k+1)$ examples. We report the mean accuracy by randomly sampling 10,000 episodes in the following experiments.

We follow the video preprocessing procedure introduced in TSN \cite{wang2016temporal}. During training we first resize each frame in the video to $256 \times 256$ and then randomly crop a $224 \times 224$ region from the video clip. For inference we change the random crop to center crop. For Kinetics dataset we randomly apply horizontal flip during training. Since the label in Something-Something V2 dataset incorporates an assumption of left and right, e.g. pulling something from left to right and pulling something from right to left, so we do not use horizontal flip for this dataset.

Following the experiment setting of CMN, we use ResNet-50 \cite{he2016deep} as the backbone network for TSN. We initialize network using pre-trained models on ImageNet \cite{deng2009imagenet}. We optimized our model with SGD \cite{bottou2010large}, with a starting learning rate of $0.001$ and decaying every 30 epochs by $0.1$. We use meta-validation set to tune the parameters, and stop the training process when the accuracy on the meta-validation set is about to decrease. We implemented the whole framework with PyTorch \cite{paszke2017automatic} framework. The whole model takes 4 TITAN Xp GPUs to train for $10$ hours.

\subsection{Evaluating Few-Shot Learning}

\begin{table}[bt]
\small
\caption{\textbf{Few-shot video classification results.} We report 5-way video classification accuracy on meta-testing set.}
\vspace{-10pt}
\begin{center}
\begin{tabular}{c|cc|cc}
\hline
       & \multicolumn{2}{c|}{Kinetics}        & \multicolumn{2}{c}{Something V2}     \\ \hline
Method & \multicolumn{1}{c|}{1-shot} & 5-shot & \multicolumn{1}{c|}{1-shot} & 5-shot \\ \hline
Matching Net \cite{zhu2018compound}  &    53.3     & 74.6 & -  & - \\ 
MAML \cite{zhu2018compound}  &    54.2     & 75.3 & -  & - \\ 
CMN \cite{zhu2018compound}  &    60.5     & 78.9 & -  & - \\ 
TSN++ & 64.5  & 77.9 & 33.6 & 43.0   \\  
CMN++ & 65.4  & 78.8 & 34.4 & 43.8  \\
TRN++ & 68.4 & 82.0 & 38.6 & 48.9 \\
TAM (ours) & \textbf{73.0}  & \textbf{85.8} & \textbf{42.8}   & \textbf{52.3}    \\
\hline 
\end{tabular}
\end{center}
\label{main_table}
\end{table}

We now evaluate the representations we learned after optimizing few-shot learning objective. We compare our method with the two following categories of baselines:

\subsubsection{Train from ImageNet Pretrained Features}

For baselines that use ImageNet pretrained features, we follow the same setting as described in CMN. As the fact that previous few-shot learning algorithms are all designed to deal with images, they usually take image-level feature encoded by some backbone network as input. To circumvent this discrepancy, we first feed frames of a video to a ResNet-50 network pretrained on ImageNet, and then average frame-level features to obtain a video-level feature. The averaged video-level feature is then served as the input of few-shot algorithms.

\noindent
\textbf{Matching Net \cite{vinyals2016matching}} We use an FCE classification layer in the original paper without fine-tuning in all experiments. The FCE module uses a bidirectional-LSTM and each training example could be viewed as an embedding of all the other examples.

\noindent
\textbf{MAML \cite{finn2017model}} Given the video-level feature as the input, we train the model following the default hyper-parameter and other settings described in \cite{finn2017model}.

\noindent
\textbf{CMN \cite{zhu2018compound}} As CMN is specially designed for few-shot video classification, it could handle video feature inputs directly. The encoded feature sequence is first fed into a multi-saliency embedding function to get a video-level feature. Final few-shot prediction is done by a compound memory structure similar to \cite{kaiser2017learning}.

For the experiment results using ImageNet pretrained backbones, we directly take the numbers from CMN \cite{zhu2018compound} to ensure fair comparison.

\begin{figure*}[ht]
    \centering
    \includegraphics[width=\textwidth]{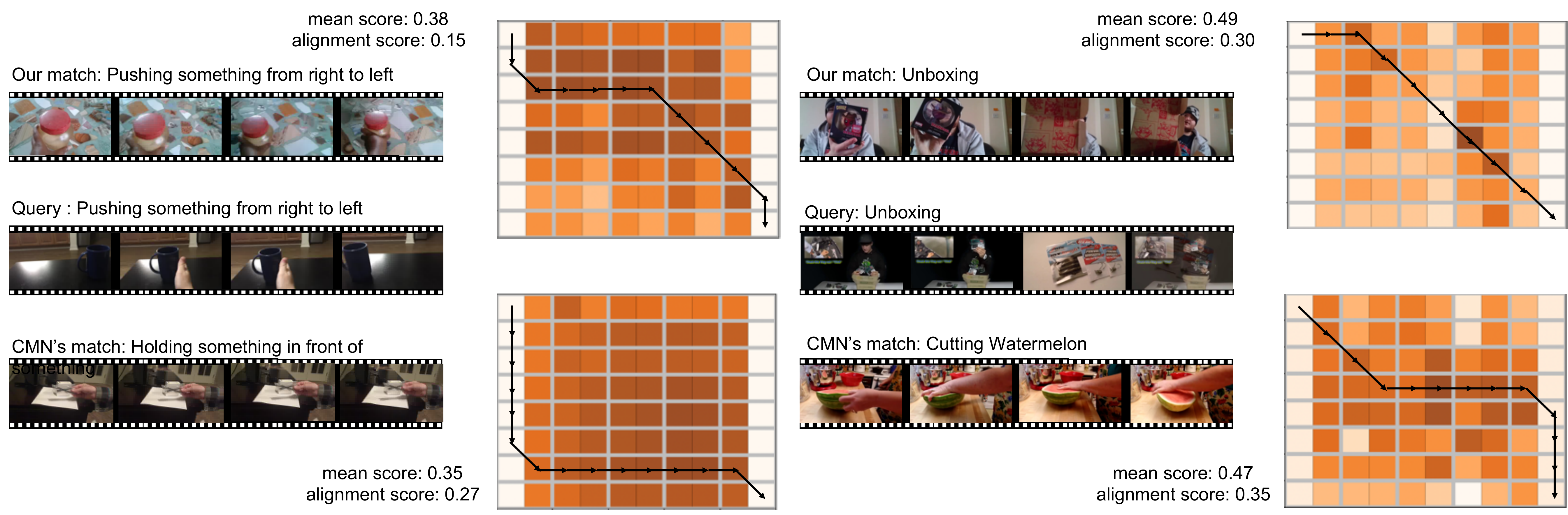}
    \caption{\textbf{Visualization of our learning results.} Comparison of our matched with CMN's matched results in an episode. Although the averaged score is quite high given the false matching and the query image, our algorithm is able to find the correct alignment path the minimize the alignment score, which ultimately results in the correct prediction.}
    \label{fig:vis}
    \vspace{-0.5cm}
\end{figure*}

\subsubsection{Finetune from Backbone on Meta Training Set}

As raised by \cite{chen19closerfewshot,gidaris2018dynamic,qi2018low}, using cosine distances between the input feature and the trainable proxy for each class could explicitly reduce intra-class variations among features during training. The rigorous experiments in \cite{chen19closerfewshot} has shown that the Baseline++ model is competitive or even surpass when compared with other few-shot learning methods. So in finetuned settings we adapt several previous approaches with the structure of Baseline++ to serve as strong baselines.

\noindent
\textbf{TSN++} For TSN++ baseline, we also use episode-based training to simulate the few-shot setting at meta-train stage to directly optimize for generalization to unseen novel classes. In order to get a video-level representation, we average over the temporal dimension of extracted per frame features for both query sets and support sets. The video level feature from support set could then serve as proxies for each novel class. We can then obtain the prediction probability for each class by normalizing these cosine distance values with a softmax function. For inference during meta-testing stage, we first forward each video in the support set to get proxies for each class. Given the proxies we can then make prediction for videos in query set.
 
\noindent
\textbf{CMN++} We follow the setting of CMN and reimplement this method by ourselves. The only difference about CMN++ and CMN is that we replace the ImageNet pretrained feature with the feature extracted by TSN++ mentioned above. 

\noindent
\textbf{TRN++} We also compare our approach against methods that attempt to learn a compact video-level representation given a sequence of image-level features. TRN \cite{zhou2018temporal} proposes a temporal relation module, which uses multilayer perceptrons (MLP) to fuse features of different frames. We refer TRN++ to one of baselines by replacing average consensus module in TSN++ with temporal relation module.

By default we conduct 5-way few-shot classification if there is no further clarification. The 1-shot and 5-shot video classification results on both the Kinetics and Something-Something V2 datasets are listed in Table \ref{main_table}. It can be concluded that our approach significantly outperforms all the baselines on both datasets. In CMN paper, the experimental observations show that fine-tuning the backbone module on the meta-training set does not improve the few-shot video classification performance. In contradiction, we find that with proper data augmentation and training strategy, a model could be trained to generalize better on unseen classes in a new domain given the meta-training set. By comparing the results of TSN++ and TRN++, we could conclude that considering temporal relation explicitly helps with model generalization on unseen classes. Compare to TSN++, the improvement brought by CMN++ is not as large as the gap on ImageNet pretrained features reported in the original paper. This may be due to the reason that we are now using a more suitable distance function (cosine distance) during meta-training so that the frame-level feature is more discriminative among unseen classes. This in turn makes it harder to improve the final prediction given those strong features as the input. Finally it is worthwhile to note that TAM outperforms all the finetuned baselines by a large margin. This demonstrates the importance of taking temporal ordering information into consideration while dealing with few-shot video classification problem.

\subsection{Qualitative Results and Visualizations}

We show qualitative results comparing CMN and TAM in Fig. \ref{fig:vis}. In particular, we observe that CMN has difficulty in differentiating two actions from different classes with very similar visual clues among all the frames, e.g., backgrounds. As can be seen from the distance matrices in Fig. \ref{fig:vis}, though our method cannot alter the fact that the two visually similar action clips will have an averagely lower frame-wise distance value, it is able to find a temporal alignment that minimize the cumulative distance score between the query action video and the true support class video while the per-frame visual clue is not evident enough. Though the mean score of TAM is lower than the match of CMN, TAM succeeds in making the right prediction via calculating a lower alignment score out of the distance matrix.

\subsection{Ablation Study}

\begin{table}[bt]
\small
\caption{\textbf{Temporal matching ablation study.} We compare our method to temporal-agnostic and temporal-aware baselines.}
\vspace{-10pt}
\begin{center}
\begin{tabular}{c|cc|cc}
\hline
       & \multicolumn{2}{c|}{Kinetics}        & \multicolumn{2}{c}{Something V2}     \\ \hline
matching type & \multicolumn{1}{c|}{1-shot} & 5-shot & \multicolumn{1}{c|}{1-shot} & 5-shot \\ \hline
Min & 52.4 & 71.6 & 29.7 & 38.5 \\
Mean & 67.8 & 78.9 & 35.2 & 45.3 \\
Diagonal & 66.2 & 79.3 & 38.3 & 48.7  \\
Plain DTW & 69.2 & 80.6 & 39.6 & 49.0 \\
TAM(Ours) & \textbf{73.0} & \textbf{85.8} & \textbf{42.8} & \textbf{52.3}  \\
\hline
\end{tabular}
\end{center}
\vspace{-0.5cm}
\label{temporal}
\end{table}

Here we perform ablation experiments to demonstrate the effectiveness of our selections of the final model. We have shown in Section 4.3 that explicitly modeling the temporal ordering plays an important role for generalization to unseen classes. We now analyze the effect of different temporal alignment approaches.

While having the cosine distance matrix $D$, there are several choices we could adopt to extract the alignment score out of the matrix, as visualized in Fig. \ref{fig:DTW}. In addition to our proposed method, we consider several heuristics for generating the scores. The first is ``Min'', where we use the minimum element in the matrix $D$ to represent the video distance value. The second is ``Mean'', for which we average over the cosine distance value of all pairs of frames. These two choices both neglect the temporal ordering. We will then introduce a few potential choices that explicitly consider sequence ordering when computing the temporal alignment score. An immediate scheme is to take an average over the diagonal of the distance matrix. The assumption made behind this approach is that the query video sequence shall be perfectly aligned with its corresponding support proxy of the same class, which could be somewhat ideal in read world applications. To allow for more adaptive alignment strategy, we introduce Plain DTW and our method. Here the Plain DTW in Table. \ref{temporal} means that there is no padding so that $W_{11}$ and $W_{TT}$ are assumed to be in the alignment path, and for each time step during computing alignment score we allow a possible movement choice among $\longrightarrow$, $\searrow$ and $\downarrow$. 

The results are shown in Table. \ref{temporal}. It can be observed that we are able to improve the few-shot learning by considering temporal ordering explicitly. There are some slight differences in performance between method Diagonal and Mean regarding to the two datasets here. There are less visual clues in each frame of Something-Something V2 than that of Kinetics, so the improvement of using Diagonal with regard to using Mean is prominent for Something-Something V2, while the gap is closed for Kinetics. However, we see that through adaptive temporal alignment, our method consistently improve the baselines on two datasets by more than 3\% accross 1-shot and 5-shots. This shows that by reinforcing the model to learn an adaptive alignment path across query videos and proxies, the final model could learn to encode better representations for the video, as well as a more accurate alignment score which could in turn help with few-shot classification.

The next ablation study is on the sensitivity of smoothing parameter $\lambda$. Previous works~\cite{chang2019d, mensch2018differentiable} have shown that using $\lambda$ empirically helps optimization in many tasks. Intuitively, a smaller $\lambda$ functions more like the min operation and a larger $\lambda$ means a heavier smoothing effect over the values in nearby positions. We experimented on $\lambda$ within the value set of $[0.01, 0.05, 0.1, 0.5, 1]$. 

\begin{figure}[t]
\centering
\includegraphics[width=1.0\linewidth]{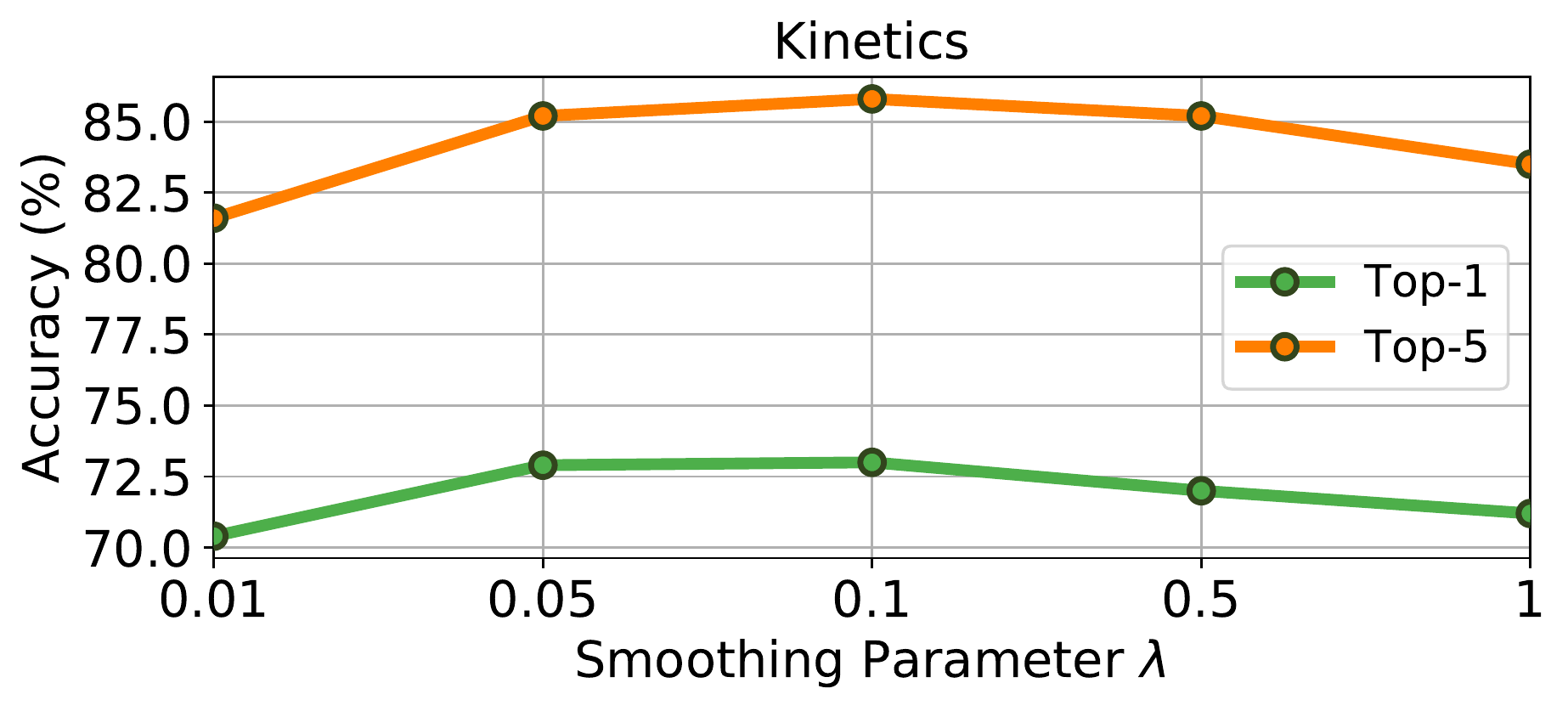}
\includegraphics[width=1.0\linewidth]{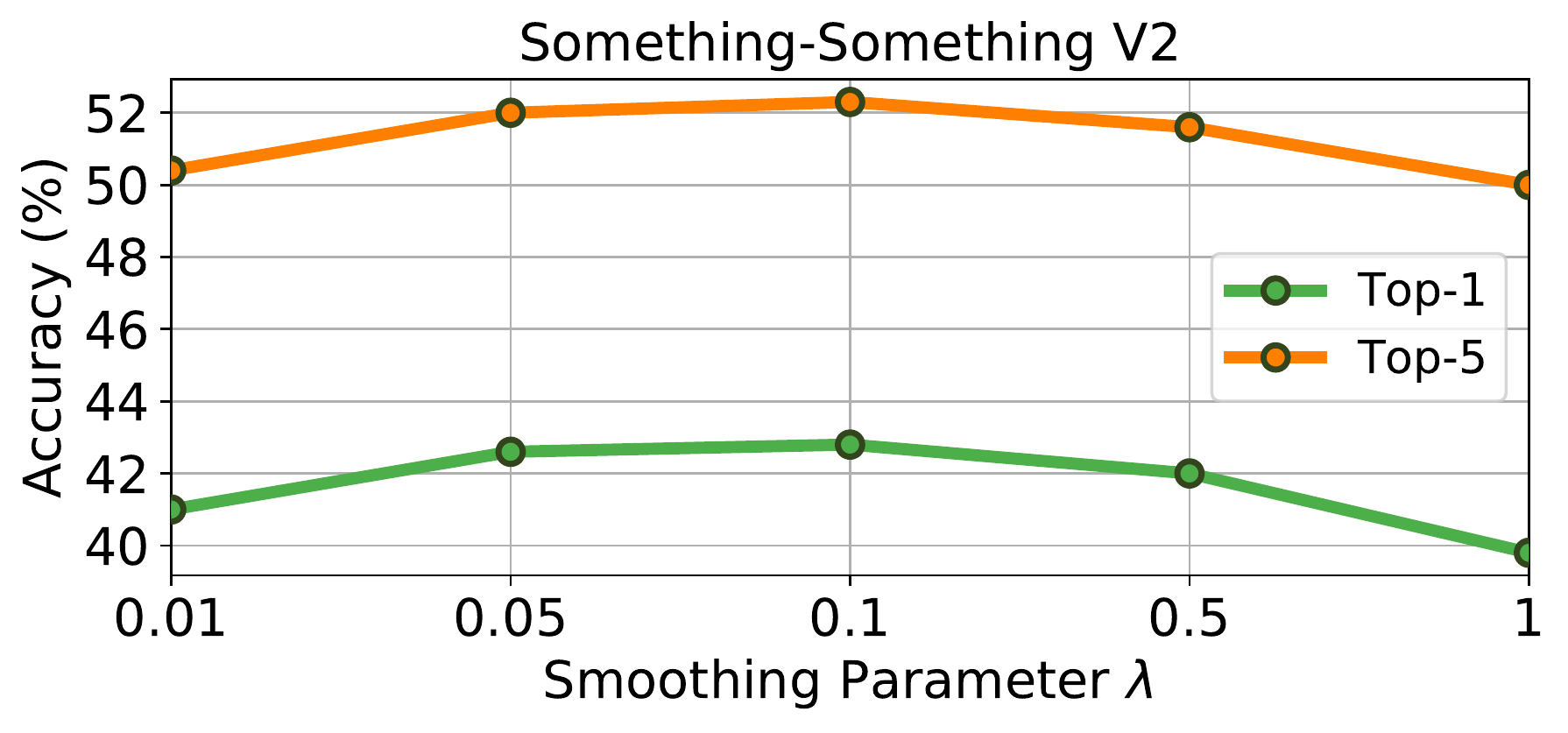}
\vspace{-2mm}
\caption{\textbf{Smoothing factor sensitivity.} We compare the effect of using different smoothing factors.}
\vspace{-5mm}
\label{fig:lambda}
\end{figure}

The results are shown in Fig. \ref{fig:lambda}. In general, the performance is stable across values of $\lambda$. We observe that in practice $\lambda$ ranges from 0.05 to 0.1 works relatively good under the setting of both two datasets. Thus we notice that a suitable $\lambda$ is essential for the representation learning. When $\lambda$ is too small, though it is able to function most similarly as the real min operator, the gradient is too imbalanced so that some pairs of frames are not adequately trained. On the contrary, a large $\lambda$ might be too smooth so that the difference among all kinds of alignments are not notable enough.

\section{Conclusion}
We propose Temporal Alignment Module (TAM), a novel few-shot framework that can explicitly learn distance measure and representation independent of non-linear temporal variations in videos using very few data. In contrast to previous works, TAM dynamically aligns two video sequences while preserving the temporal ordering and it further uses continuous relaxation to directly optimize for the few-shot learning objective in an end-to-end fashion. Our results and ablations show that our model significantly outperforms a wide range of competitive baselines and achieves state-of-the-art results on two challenging real-world datasets. 

\paragraph{Acknowledgements} This work has been partially supported by JD.com American Technologies Corporation (”JD”) under the SAILJD AI Research Initiative. This article solely reflects the opinions and conclusions of its authors and not JD or any entity associated with JD.com.

{\small
\bibliographystyle{ieee}
\bibliography{main}
}

\end{document}